\documentclass{article}

    \PassOptionsToPackage{numbers, compress}{natbib}

    \usepackage[preprint]{neurips_2025}

\usepackage[numbers]{natbib}
\usepackage[utf8]{inputenc} 
\usepackage[T1]{fontenc}    
\usepackage{hyperref}       
\usepackage{url}            
\usepackage{booktabs}       
\usepackage{graphicx}       
\usepackage{amsfonts}       
\usepackage{nicefrac}       
\usepackage{microtype}      
\usepackage{xcolor}         
\usepackage{booktabs}
\usepackage[normalem]{ulem}
\useunder{\uline}{\ul}{}

\usepackage{physics}
\usepackage{amsmath}
\usepackage{tikz}
\usepackage{mathdots}
\usepackage{yhmath}
\usepackage{cancel}
\usepackage{color}
\usepackage{siunitx}
\usepackage{array}
\usepackage{multirow}
\usepackage{amssymb}
\usepackage{gensymb}
\usepackage{tabularx}
\usepackage{extarrows}
\usepackage{booktabs}
\usetikzlibrary{fadings}
\usetikzlibrary{patterns}
\usetikzlibrary{shadows.blur}
\usetikzlibrary{shapes}

\title{VM14K: First Vietnamese Medical Benchmark}

\author{
  Thong Nguyen\thanks{These authors contributed equally to this work.}\\
  Vietnam National University \\
  \And
  Duc Nguyen\footnotemark[1] \\
  Dickinson College \\
  \And
  Minh Dang\footnotemark[1] \\
  Columbia University \\
  \And
  Thai Dao\footnotemark[1] \\
  Venera AI \\
  \And
  Long Nguyen \\ 
  Carnegie Mellon University \\
  \And
  Quan H. Nguyen \\ 
  University of Maryland \\
  \And
  Dat Nguyen \\ 
  Venera AI \\
  \And
  Kien Tran \\ 
  Venera AI \\
  \And
  Minh Tran \\ 
  Foreign Trade University\\
  }

\begin{document}
\maketitle
\begin{abstract}

Medical benchmarks are indispensable for evaluating the capabilities of language models in healthcare for non-English-speaking communities,therefore help ensuring the quality of real-life applications. However, not every community has sufficient resources and standardized methods to effectively build and design such benchmark, and available non-English medical data is normally fragmented and difficult to verify. We developed an approach to tackle this problem and applied it to create the first Vietnamese medical question benchmark, featuring 14,000 multiple-choice questions across 34 medical specialties. Our benchmark was constructed using various verifiable sources, including carefully curated medical exams and clinical records, and eventually annotated by medical experts. The benchmark includes four difficulty levels, ranging from foundational biological knowledge commonly found in textbooks to typical clinical case studies that require advanced reasoning. This design enables assessment of both the breadth and depth of language models' medical understanding in the target language thanks to its extensive coverage and in-depth subject-specific expertise. We release the benchmark in three parts: a sample public set (4k questions), a full public set (10k questions), and a private set (2k questions) used for leaderboard evaluation. Each set contains all medical subfields and difficulty levels. Our approach is scalable to other languages, and we open-source our data construction pipeline to support the development of future multilingual benchmarks in the medical domain. \href{https://venera-ai.github.io/VM14K/}{Project Page.}
  
\end{abstract}

\section{Introduction} 

The rise of Large Language Models (LLMs) in healthcare has become more prominent than ever, projecting tremendous impacts across various fields such as clinical research, diagnosis support systems and other different medical specialties \cite{survey2024, survey2025, stanford_hai_medhelm}. In such a critical field like medicine, medical benchmarks are indispensable for evaluating the quality of these models before applying to real-world application. However, the lack of diversity in terms of language - most medical resources including data, models, and benchmarks are primarily in English, or  Chinese - implies the need for developing benchmarks in other non-popular languages to support local communities. \cite{jin2023betteraskenglishcrosslingual, kasai2023evaluatinggpt4chatgptjapanese, kweon2024kormedmcqamultichoicequestionanswering, yu2024tcmdtraditionalchinesemedicine}. Force prompting in target language or pure translate question could be an option, however it might suffer from information loss, hallucination due to inherent difference in language semantic  \cite{gérardin2023impacttranslationbiomedicalinformation, jin2023betteraskenglishcrosslingual}. Additionally, while it is straightforward to develop and test LLMs on universal benchmarks, testing for highly-specialized fields like medical in local languages requires careful design with detailed standards \cite{jin2019pubmedqa, LiveMedQA2017, hendryckstest2021, pmlr-v174-pal22a}. Several reports highlight how off-the-shelf models and benchmarks are not optimal for multilingual applications; common problems include missing concepts in specific languages or special characteristics in the medical language of particular regions \cite{gérardin2023impacttranslationbiomedicalinformation, jin2023betteraskenglishcrosslingual}.

Based on our knowledge, there is no universal process for designing local medical benchmarks for LLMs. Collection methods are often fragmented or even nonexistent in many languages, and the definition and level of detail sometimes differ across regions, posing difficulties in the testing process. Moreover, it poses significant challenges such as data verification, as medical data is normally hard to verify and  heavily skewed toward common diseases in the distribution of questions\cite{wang2025safetychallengesaimedicine, qin2025opportunities, park2024assessing}.

Virtually all countries have a corpus of medical textbooks and testing systems for their students and doctors. Even if not every community has a modernized medical education system, these materials are at least verified or standardized by experts to maintain educational quality. We consider this a valuable source of truth that is already verified by experts in the field and continuously maintained and updated. More importantly, most historical data is public and could be explored from different perspectives to serve the purposes of LLMs in the medical domain, from training to testing model quality \cite{zhang2024critical, ir.2024.27, qiu2024building, 9184044}.

To advance the progress of local LLMs in the medical field for non-English speaking communities, we \textbf{(1) developed a scalable framework} that curate medical data from various sources (websites, text, PDF), \textbf{(2) defined a simple yet flexible standard} for designing medical benchmarks for local LLMs, and \textbf{(3) created a first medical benchmark for Vietnamese} with 14,000 questions spanning 34 medical specialties, carefully annotated by medical students across four difficulty levels. We evaluated robustness of our benchmark on multiple foundation and open source medical models. We also open-source our data infrastructure to contribute to advancements in the medical LLM field with broader impact. 

\section{Related works} 
\subsection{Non-English medical benchmark} 
Existing non-English benchmarks are still fragmented and limited in scope. For example, Chinese and Korean medical benchmarks have shown their evaluation feasibility \cite{liu2023benchmarkinglargelanguagemodels, kweon2024kormedmcqamultichoicequestionanswering,  yu2024tcmdtraditionalchinesemedicine, tian-etal-2019-chimed}, but they lack a standardized structure for difficult levels, question formats, and annotation guidelines. Furthermore, they also limit themselves to a small range of specialties only. Our benchmark directly address these drawbacks by covering across 34 medical topics with a predefined schema.  

Due to the lack of native language resources, some researches have tried to translate English benchmarks into other languages \cite{singhal2022largelanguagemodelsencode, kasai2023evaluatinggpt4chatgptjapanese, qiu2024building}. However, it often introduces terminology inconsistency, ambiguity, or cultural misalignment, which can reduce the quality of training and evaluation process. Even high-quality machine translations can lead to clinical nuance and  misinformed content in the biomedical domain \cite{gérardin2023impacttranslationbiomedicalinformation}. Furthermore, medical education is different between regions and may not align with local structure or practice standards \cite{jin2023betteraskenglishcrosslingual}. 

In many regions, including Southeast Asia, there is no standardized effort to benchmark local-language medical knowledge; rather than relying on datasets from medical licensing exams or student materials, there is no framework to support cross-specialty generalization and scalability \cite{ir.2024.27}. Our work addresses this gap by proposing a systematic framework, firmly based on local medical education resources, to ensure high quality and relevance. 

\subsection{Medical data curation} 
For medical QA, medical benchmarks are majorly curated from English resources such as USMLE practice questions, PubMed research abstracts (MMLU Medical \cite{hendryckstest2021} \cite{hendrycks2021ethics}, MedQA \cite{jin2020disease}, PubMedQA\cite{jin2019pubmedqa}), U.S. National Library of Medicine(TREC-2017 LiveQA \cite{LiveMedQA2017}). Occasionally, there are datasets collected from other non-English resources such as medical board exam in China (cMedQA2 \cite{8548603}), AIIMS and NEET PG from India (MedMCQA\cite{pmlr-v174-pal22a}). Besides, there exists datasets that were synthesized from other non-mainstream resources such as Reddit, StackExchange (BiQA \cite{9184044}). These datasets usually collected and curated from resources that have the same structure of data or a subspace of medical knowledge space, which is not often verified and representative enough to be used for building and evaluating medical LLMs. In addition, manually synthesizing from different resources is not scalable and efficient due to the heterogeneous structure of data from different resources. Our dataset is collected and processed through a scalable pipeline that gather online Vietnamese medical resources (medical textbook, medical exams, online medical quizzes \& flashcards). Based on these verifiable sources, selected data was carefully filtered, annotated to build a complete benchmark. This pipeline is applicable to any languages, allowing researchers to conduct data synthesizing on their desire language for pre-training, fine-tuning and downstream tasks.

\section{Methods} 
\subsection{Benchmark design} 

Among several existing methodologies for benchmark development, we designed our approach with the primary criteria of medical comprehensiveness, as the benchmark must be broad enough to cover the full spectrum of medical knowledge while being clear, replicable, and extendable to other languages. The selection of medical specialties was intentionally comprehensive to ensure both breadth across the medical field and sufficient depth within each specialty. Our benchmark includes 34 distinct medical categories, carefully selected to provide a holistic evaluation of medical knowledge. This extensive categorization \ref{tab:category} serves multiple critical purposes: 

\textbf{(1) Complete Coverage of Medical Practice:} The categories span the entire medical field, from primary care to highly specialized domains, ensuring that no significant area of medical practice is overlooked.

\textbf{(2) Balanced Representation:} We deliberately included both common specialties (like Internal Medicine and Pediatrics) and more specialized fields (such as Nuclear Medicine and Palliative Care) to prevent bias toward frequently encountered medical concepts.

\textbf{(3) Integrated Medical Knowledge: }The selected categories not only represent universally recognized medical specialties practiced across different healthcare systems worldwide, but also have categories related to traditional medical systems that remain relevant in many communities, especially Eastern. This design making our framework adaptable to various cultural and linguistic contexts.

\textbf{(4) Public Health and Preventive Focus:} 
Categories such as Public Health and Preventive Healthcare were selected to ensure that the benchmark evaluates the understanding of population-level interventions and disease prevention strategies, not just treatment-oriented knowledge. By including questions on disease surveillance, vaccination strategies, health education campaigns, and environmental risk mitigation, our benchmark captures the full spectrum of understanding needed to anticipate and avert outbreaks, reduce health disparities, and promote social well-being. The benchmark emphasizes the importance of upstream interventions, such as policy development, community engagement, and behavior change frameworks, that can prevent illness before it occurs, complementing treatment-focused competencies and fostering a truly holistic approach to healthcare.

To comprehensively assess the capabilities of medical LLMs, we introduce a four-tier difficulty framework, ranging from fundamental knowledge to complex clinical reasoning. This framework is designed to evaluate LLMs across a spectrum of cognitive skills necessary for medical professionals. This definition of levels \ref{tab:diff_lv}  offered several advantages:

\textbf{(1) Comprehensive Evaluation:} It allows for a comprehensive assessment of LLMs by covering a wide range of cognitive skills, from basic recall to complex reasoning. It aligns with the cognitive skills expected of medical professionals, mirroring the progression from foundational knowledge to advanced clinical practice. This ensures that the evaluation is not limited to a single aspect of medical knowledge.

\textbf{(2) Nuanced Assessment:} By differentiating between levels, the framework provides a nuanced understanding of an LLM's strengths and weaknesses. For instance, an LLM might excel at recalling factual information but struggle with complex clinical scenarios.

\textbf{(3) Alignment with Medical Expertise:} The framework aligns with the cognitive skills expected of medical professionals, mirroring the progression from foundational knowledge to advanced clinical practice. The "Challenging" and "Hard" levels assess skills crucial for real-world medical practice, such as diagnostic reasoning and clinical decision-making, guiding the development of more effective training strategies and architectures.

\begin{figure}
    \centering
    \includegraphics[width=0.75\linewidth]{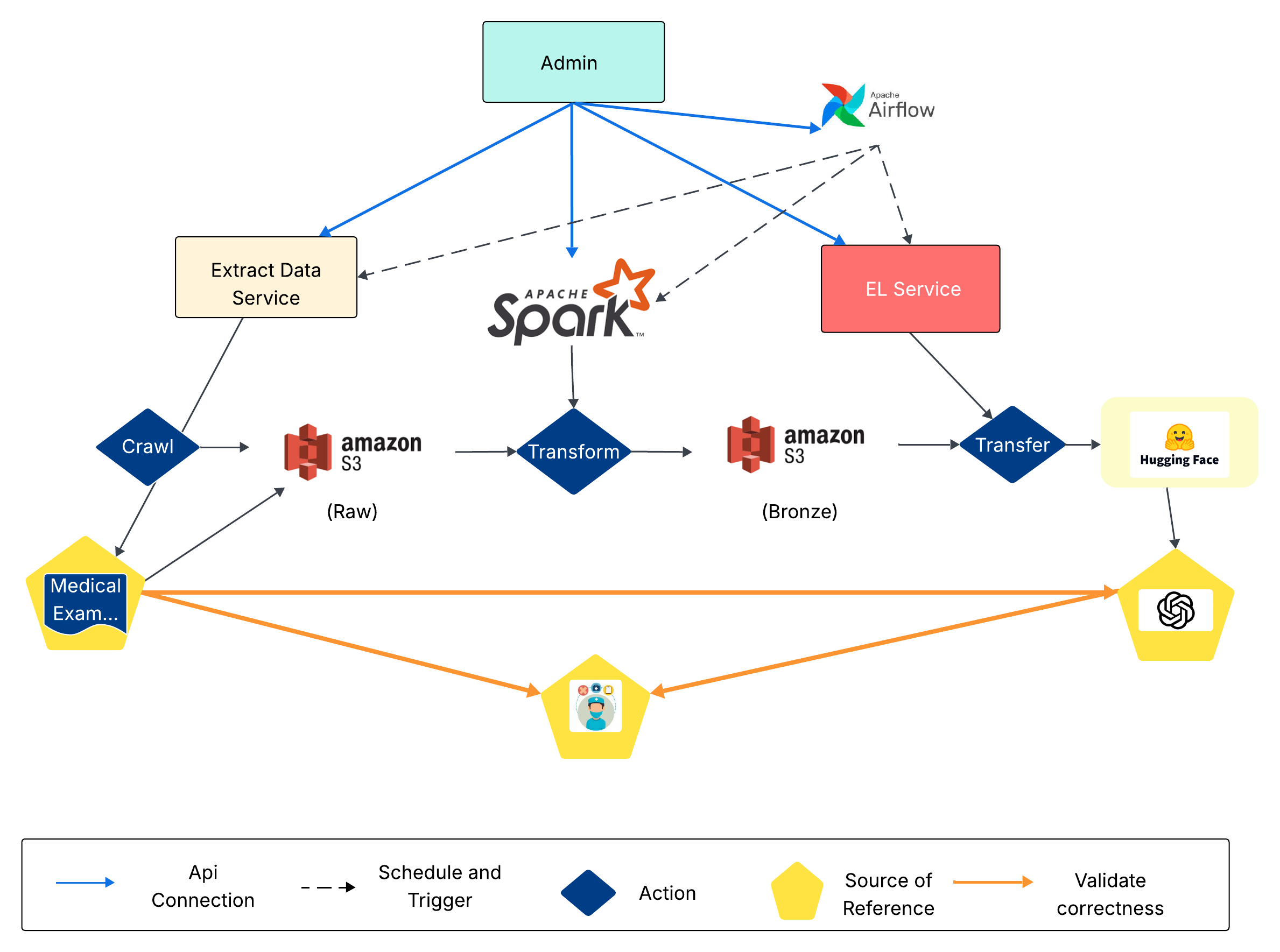}
    \caption{Data curation and verification pipeline.}
    \label{fig:pipeline}
\end{figure}

\subsection{Data collection}  

\subsubsection{Data pipeline}

\paragraph{Extract data} The crawler service is the core component our systems, this service is implemented in Python to extract medical data from reliable resources including medical textbooks, medical exams/tests from medical institutions, and public medical records from research institutions. The extracted data are stored in Amazon S3 (AWS) in a folder in its raw format including text, document files, PDF files, images, along with metadata.

\paragraph{Transformation} We leveraged Spark to efficiently transform large amounts of data in parallel. This transformation process standardized most of the text data to JSON format while image and PDF data were transferred to another flow. The transformed data are stored in S3 in the destination path, which we refer to as the bronze zone. Further advanced transformed data can be stored in the silver zone or the gold zone depending on the storage structure convention.

\paragraph{Transfer data} We implemented an EL (Extract-Load) service from scratch, which helps transfer data from various source types to various destination types based on needs. This design allowed us to craft data effectively for different purposes besides benchmarking.

\paragraph{Orchestral workflow} We used Apache Airflow for orchestration as our services would be triggered periodically to update and process new incoming data.

\paragraph{Incremental extracting} We use PostgreSQL as a metadata database for data tracking to handle data duplication. We concatenate and hash strings created from fields to determine whether the data is new or not. The hash column serves as a unique identifier in the metadata. The crawler service uses metadata to detect which data have been extracted and to avoid duplicates.



\paragraph{Data processing} 
The data was reformatted by removing extra index, HTML tags and then standardized to our JSON format while PDFs and documents were further extracted and deduplicated as needed.

\subsubsection{Structured data processing} 
We applied the method below for PDF and Microsoft Word files to extract and standardize questions from these sources.

\paragraph{Extract content} We used docx2python~\footnote{https://github.com/ShayHill/docx2python} to extract text from Microsoft Words and pymupdf~\footnote{https://github.com/pymupdf/PyMuPDF} to extract text from PDF. The raw text of questions (mostly multiple-choice questions and their options) is saved in the same output file.

\paragraph{Extract questions}
We utilized OpenAI's GPT-4o\cite{gpt4o} and Gemini 2.0 Flash\cite{gemini20} to extract questions to structured output \ref{tab:tags}. Since our full raw dataset included 100,000 questions and produced large text files, we divided theses into chunks for parallel batch processing. We noticed that each question usually has 5 to 7 lines of text, so we set the chunk size to 100 lines (to ensure within the context length limit of GPT-4o) and overlap 5 lines to prevent losing questions. For the questions that do not have difficultLevel and medicalTopic, we used GPT-4o and Geimini Flash to label those based on our definition \ref{tab:category}, \ref{tab:diff_lv}.

The extracted questions contain the following fields:
\begin{table}[h!]
\centering
\caption{Extracted question field}
\label{tab:question_attributes}
\begin{tabular}{p{0.2\textwidth}p{0.6\textwidth}}
\toprule
\textbf{Field} & \textbf{Description} \\
\midrule
question & Medical multiple-choice question \\
answers & List of options for the answer \\
regularFormat & True if the question is multiple-choice. False otherwise \\
correctOption & The correct answer according to GPT-4o \\
difficultLevel & Level of difficulty of the question evaluated by GPT-4o \\
medicalTopic & List of relevant medical topics \\
\bottomrule
\end{tabular}
\label{tab:tags}
\end{table}


\paragraph{Data verification}

\begin{figure}
    \centering
    \includegraphics[width=0.75\linewidth]{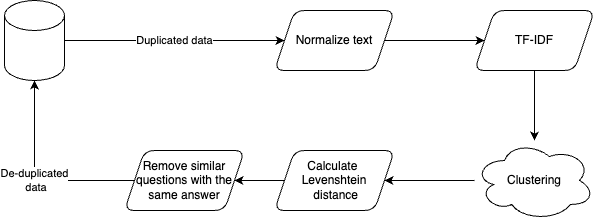}
    \caption{Duplication removal process}
    \label{fig:deduplication_diagram}
\end{figure}

Figure \ref{fig:deduplication_diagram} describes the pipeline that we built to remove redundant entries. This approach is effective for the Vietnamese language, where accents create many variances. Additionally, removal based on clustering and the Levenshtein distance is suitable when the questions and answers are slightly different at the character level, as is common in chemical-related topics. Afterwards, the answer verification step includes three sources of references: (1) the correct answers from the source data, (2) the answers from foundation LLMs, and (3) the answers from medical experts. We ranked questions for annotation based on mutual disagreement and level of difficulty. The first questions to be verified by human experts are the easiest ones, where the original source answers and the answers from foundation models differ. This approach is based on the assumption that all sources of truth should have mutual agreement on easy questions. Since the annotation process took a long time and the data distribution was heavily skewed toward common topics, we sampled 14,000 questions from a total of 100,000, ensuring an even representation of the 34 medical categories included in our benchmark. This prevents common topics from dominating the dataset and allows for a more balanced evaluation of LLM performance across different areas of medical knowledge. This whole process offers early detection of potential problems, ensures representativeness, and provides a holistic verification of our benchmark.

\section{Experiments} 

In this section, we conduct a comprehensive evaluation of several LLMs including general models with multilingual capability and medical specialized models. We employ two evaluation metrics: pass@$k$ (with $k = 3$, $k = 1$ for non-reasoning models and $k = 1$ for reasoning models, except for Deepseeek-R1) and ensemble for all models. The ensemble setup involved running the model three times on the same set of questions, but choices were shuffled independently for each run, then voting is used for final aggregation. We analyze the results and provide insight about the potential and limitation of those LLMs in Vietnamese medical domain. The main results are shown in Table \ref{tab:pass_at} and Table \ref{tab:ensemble}.

\begin{table}[!ht]
\centering
\caption{Model performance on VM14K with pass@$k$ metric. Values are accuracies in percentages. Challenging and Hard questions are combined under "Hard". }
\label{tab:pass_at}
\vspace{5pt}
\resizebox{0.95\linewidth}{!}{
\begin{tabular}{p{0.32\linewidth}cccccccc}
\toprule
& \textbf{Overall} & \textbf{Easy} & \textbf{Medium} & \textbf{Hard} & \textbf{Overall} & \textbf{Easy} & \textbf{Medium} & \textbf{Hard} \\ 
\cmidrule(r){2-5} \cmidrule(r){6-9}
\textbf{Models}              & \multicolumn{4}{c}{pass@3}        & \multicolumn{4}{c}{pass@1}        \\ 
\midrule
\multicolumn{9}{c}{Closed-source models} \\ 
\midrule
GPT-4o\cite{gpt4o}              & \textbf{80.40} & \textbf{82.78} & \textbf{79.57} & \textbf{77.95} & 72.74 & 75.11 & 72.06 & 69.19 \\ 
o3-mini\cite{o3mini}             & -     & -     & -     & -     & 71.42 & 73.21 & 70.92 & 68.58 \\ 
Claude 3.5 Sonnet\cite{claude35sonnet}   & 75.62 & 77.50 & 74.81 & 74.43 & 71.46 & 73.12 & 70.77 & 70.58 \\ 
Gemini 2.0 Flash\cite{gemini20}    & 77.92 & 80.16 & 77.34 & 74.77 & \textbf{75.67} & \textbf{78.03} & \textbf{75.18} & \textbf{71.75} \\ 
\midrule
\multicolumn{9}{c}{Open-source models} \\ 
\midrule
Deepseek-R1\cite{dsr1}                     & \textbf{83.02} & \textbf{84.68} & \textbf{82.32} & \textbf{82.31} & \textbf{78.17} & \textbf{79.89} & \textbf{77.54} & \textbf{77.03} \\ 
Qwen3-32B\cite{yang2025qwen3technicalreport}                       & -     & -     & -     & -     & 72.47 & 74.37 & 72.10 & 68.79 \\ 
Qwen3-30B-A3B                   & -     & -     & -     & -     & 71.32 & 73.69 & 70.28 & 65.08 \\ 
Llama 4 Maverick\cite{llama4}                & 73.13 & 74.68 & 72.66 & 71.17 & 71.76 & 72.95 & 71.54 & 69.66 \\ 
Phi-4\cite{abdin2024phi4technicalreport}                           & 69.18 & 71.57 & 68.77 & 60.99 & 51.15 & 53.82 & 50.44 & 45.52 \\ 
Gemma3-27B-Instruct\cite{gemma3}             & 64.45 & 67.36 & 63.65 & 55.65 & 63.38 & 66.46 & 62.50 & 53.99 \\ 
Gemma3-12B-Instruct             & 60.03 & 62.86 & 59.13 & 53.16 & 58.05 & 60.60 & 57.39 & 50.84 \\ 
Llama-3.1-8B-Instruct\cite{grattafiori2024llama3herdmodels}           & 60.88 & 63.55 & 59.88 & 55.28 & 48.73 & 52.33 & 47.36 & 42.08 \\ 
\midrule
\multicolumn{9}{c}{Open-source medical specialized models} \\ 
\midrule
Llama-3-70B-UltraMedical\cite{zhang2024ultramedical}   & \textbf{67.34} & \textbf{70.53} & \textbf{65.64} & \textbf{65.07} & \textbf{54.96} & \textbf{58.34} & \textbf{53.55} & 51.07 \\ 
Meditron3-70B\cite{chen2023meditron70bscalingmedicalpretraining}              & 59.55 & 62.86 & 58.17 & 45.42 & 42.27 & 45.57 & 40.86 & 37.17 \\ 
HuatuoGPT-o1-8B\cite{chen2024huatuogpto1medicalcomplexreasoning}            & -     & -     & -     & -     & 54.52 & 57.93 & 53.14 & \textbf{52.19} \\ 
Llama-3.1-8B-UltraMedical   & 48.01 & 52.26 & 46.85 & 38.36 & 41.86 & 46.13 & 40.66 & 31.06 \\ 
Meditron3-8B                & 30.04 & 31.25 & 28.82 & 29.54 & 23.05 & 23.81 & 22.08 & 23.20 \\ 
\bottomrule
\end{tabular}
} 
\end{table}
\begin{table}[!ht]
\centering
\caption{Model performance on VM14K with ensemble metric. Values are accuracies in percentages. Challenging and Hard questions are combined under "Hard". *We included Pass@$k$ column for reference.}
\label{tab:ensemble}
\vspace{5pt}
\resizebox{0.7\linewidth}{!}{
\begin{tabular}{p{0.3\linewidth}ccccc}
\toprule
\textbf{Models} & \textbf{Pass@\textit{k}}* & \textbf{Overall} & \textbf{Easy} & \textbf{Medium} & \textbf{Hard} \\ 
\midrule
\multicolumn{6}{c}{Closed-source models} \\ 
\midrule
GPT-4o              & \textbf{80.40} & 73.80 (-6.60) & 76.36 & 72.79 & 71.87 \\ 
GPT-o3-mini         & 71.42 & 72.02 (+0.60) & 74.11 & 71.35 & 69.33 \\ 
Claude 3.5 Sonnet   & 75.62 & 72.31 (-3.31) & 73.81 & 71.71 & 71.67 \\ 
Gemin 2.0 Flash     & 77.29 & \textbf{76.30 (-0.99)} & \textbf{78.89} & \textbf{75.58} & \textbf{72.84} \\ 
\midrule
\multicolumn{6}{c}{Open-source models} \\ 
\midrule
Deepseek-R1                     & \textbf{83.02} & 78.42 (-4.60) & 80.59 & 77.60 & 76.78 \\ 
Qwen3-32B                       & 72.47 & \textbf{84.96 (+12.49)} & \textbf{86.60} & \textbf{84.51} & \textbf{82.23} \\ 
Qwen3-30B-A3B                   & 71.32 & 84.04 (+12.72) & 85.48 & 83.63 & 81.42 \\ 
Llama 4 Maverick                & 73.13 & 72.46 (-1.33) & 73.71 & 72.06 & 71.42 \\ 
Phi-4                           & 69.18 & 72.25 (+3.07) & 74.59 & 71.89 & 67.25 \\ 
Gemma3-27B-Instruct             & 64.45 & 77.11 (+12.66) & 79.16 & 76.33 & 72.64 \\ 
Gemma3-12B-Instruct             & 60.03 & 74.09 (+14.06) & 76.12 & 73.57 & 67.68 \\ 
Llama-3-8B-Instruct             & 60.88 & 69.92 (+9.04) & 72.57 & 68.84 & 68.06 \\ 
\midrule
\multicolumn{6}{c}{Open-source medical specialized models} \\ 
\midrule
Llama-3-70B-UltraMedical    & \textbf{67.34} & 72.57 (+5.23) & 75.10 & 71.30 & 72.33 \\ 
Meditron3-70B               & 59.55 & 56.10 (-3.45) & 59.31 & 54.98 & 45.42 \\ 
HuatuoGPT-o1-8B             & 54.52 & \textbf{73.89 (+19.37)} & \textbf{76.41} & \textbf{72.99} & \textbf{73.03} \\ 
Llama-3.1-8B-UltraMedical   & 48.01 & 56.24 (+8.23) & 60.19 & 55.45 & 44.27 \\ 
Meditron3-8B                & 30.04 & 30.09 (+0.05) & 31.69 & 28.80 & 28.57 \\ 
\bottomrule
\end{tabular}
} 
\end{table}

\subsection{General model with multilingual capability}







\paragraph{Results} As shown in Table \ref{tab:pass_at}, Deepseek R1 outperforms all other models significantly in both Pass@1 and Ensemble. Apart from  Deepseek R1 outstanding performance, other reasoning models (o3-mini, Qwen 3) are not better than non-reasoning models such as Claude 3.5 Sonnet, GPT-4o, Gemini 2.0 Flash. This is in direct conflict with the performance on MedQA\cite{jin2020disease}, an English medical benchmark, where o3-mini and Qwen 3 performed better than or at the same level as Deepseek R1 and much better than non-reasoning models\cite{valsai}. These observations suggest that for our benchmark, other factors, such as Vietnamese capability, are better predictors of performance than medical ability in English.

The ensemble method with shuffled choices shows an improvement across the board for all models. GPT-4o saw the highest improvement in ensemble performance over Pass@1. Interestingly, the lowest improvement from ensembling is observed for Deepseek R1, the best-performing model.


\begin{figure}
    \centering
    \includegraphics[width=0.75\linewidth]{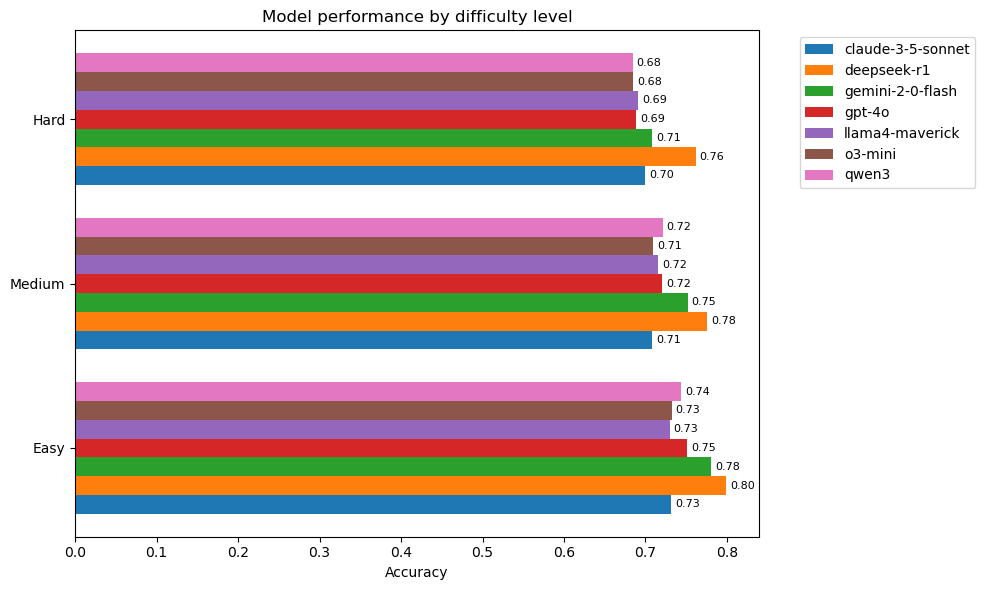}
    \caption{Performance of general models on the benchmark by difficulty levels. Challenging and Hard questions are combined under "Hard".}
    \label{fig:diff_lvl_grouped_barh}
\end{figure}

\paragraph{Difficulty Levels} As shown in Figure \ref{fig:diff_lvl_grouped_barh}, the performance of most models decreases as difficulty increases. However, the difference in accuracy between difficulty levels is not as large as in smaller open-source models. In particular, Claude 3.5 Sonnet and Llama 4 Maverick maintain relatively consistent accuracy across difficulty levels, although there is still a slight downward trend.

\paragraph{Medical Topics} Table \ref{tab:accuracy-by-medical-topic} shows that these foundation models have different levels of accuracy for different medical topics. Among the top 10 most popular medical topics, Oncology and Endocrinology are topics with higher accuracy while most models struggle with Pediatrics. These differences in performance are distributed quite evenly among foundation models, i.e. there are no clear signs any models have strengths in some certain topics and weaknesses in others.

\begin{table}[]
\centering
\caption{Pass@1 by Top 10 popular medical topics. Values are accuracies in percentages.}
\vspace{5pt}
\resizebox{0.9\textwidth}{!}{
\begin{tabular}{p{0.15\linewidth}ccccccc}
\toprule{}
\textbf{Medical Topic}   & \textbf{GPT} & \textbf{o3} & \textbf{Llama 4} & \textbf{Claude 3.5} & \textbf{Qwen3} & \textbf{Gemini} & \textbf{Deepseek} \\
& \textbf{4o} & \textbf{mini} & \textbf{Maverick} & \textbf{Sonnet} & \textbf{32B} & \textbf{2.0 Flash} & \textbf{R1} \\
\midrule{}
Gastroenterology          & 72.07 & 73.19 & 73.31 & 72.43 & 74.02 & 77.01 & \textbf{81.47} \\
Obstetrics and Gynecology & 74.61 & 69.80 & 72.29 & 74.12 & 71.55 & 77.18 & \textbf{79.71} \\
Pulmonology               & 70.87 & 70.82 & 68.12 & 68.66 & 70.23 & 74.00 & \textbf{77.35} \\
Infectious Diseases       & 73.23 & 72.24 & 71.74 & 71.63 & 71.74 & 76.20 & \textbf{77.79} \\
Endocrinology             & 76.92 & 76.79 & 76.34 & 75.63 & 78.01 & 78.27 & \textbf{81.24} \\
Oncology                  & 75.86 & 75.74 & 76.58 & 75.15 & 77.41 & 79.67 & \textbf{82.64} \\
Pathology                 & 71.23 & 67.61 & 69.16 & 70.97 & 73.03 & 74.19 & \textbf{78.19} \\
Pediatrics                & 67.74 & 67.48 & 64.52 & 68.65 & 67.23 & 68.39 & \textbf{73.16} \\
Radiology                 & 70.81 & 72.05 & 69.57 & 69.25 & 70.19 & 74.84 & \textbf{77.48} \\
Surgery                   & 68.77 & 66.83 & 68.77 & 68.61 & 68.77 & 71.36 & \textbf{75.89} \\
\bottomrule{}
\end{tabular}
}
\label{tab:accuracy-by-medical-topic}
\end{table}

\begin{figure}
    \centering
    \includegraphics[width=0.85\linewidth]{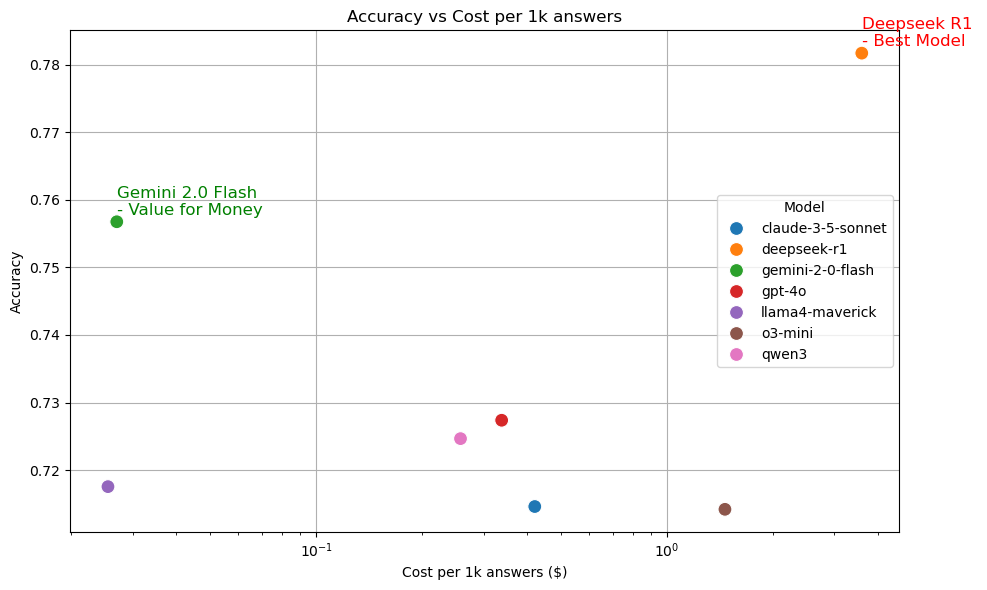}
    \caption{Cost versus performance analysis}
    \label{fig:cost-performance-analysis}
\end{figure}

\paragraph{Cost-Performance analysis} Figure \ref{fig:cost-performance-analysis} shows the cost analysis of the foundation models, which can inform the use of these models as a service for Vietnamese medical use cases. Deepseek R1 is the best model at the cost of \$ 3.59 inference cost to answers 1000 questions, in contrast with the other reasoning model, OpenAI's o3-mini, which is the second-most expensive model but also the most inaccurate. Gemini 2.0 Flash is the best value-for-money model, only trailing Deepseek R1 in performance and is the cheapest among the 7 models.

\subsection{Medical Specialized Models} 



\paragraph{Results} In general, larger models (70B) have more capacity to learn complex medical knowledge and reasoning patterns, leading to higher accuracy. This trend is clear across all difficulty levels. They can also maintain their performance better as question difficulty increases, which means scale helps with complex, multi-step reasoning required for harder questions. The drop in accuracy from Easy to Hard is less severe for larger models, except for HuatuoGPT-o1-8B: despite being 8B, it performs at the top, likely due to specialized training or data focused on medical reasoning. However, the rest of the 8B models are consistently the lowest performers; and the best results are achieved when both scale and specialization are present (for example Llama-3-70B-UltraMedical).

\paragraph{Ensemble} The ensemble results are higher than pass@1 and often close to or better than pass@3, showing that combining multiple runs or shuffles helps models "cover" more correct answers, especially for difficult questions. We also observe a large shift in performance between Llama-3-70B-UltraMedical and HuatonGPT-o1-8B as the evaluation method changed from Pass@\textit{k} to Ensemble where the latter is the better. We argue that the Ensemble metric is more holistic to evaluate of the model as some randomness can easily affect Pass@\textit{k}. That said, a small well-trained reasoning model model could reach similar performance of a larger one.

Also, it is important to note that the number of open-source medical language models is relatively low, and most of the publicly available ones are small-scale. The majority of large-scale medical language models are currently proprietary. This is understandable, since training medical LLMs is costly and requires careful alignment and adherence to regulations.



\section{Conclusion} 

In this work, we address the gap in linguistic diversity for medical benchmark by introducing an complete approach to collect, process reliable data sources and design benchmark. We applied this method to build the first Vietnamese medical benchmark (14,000 questions, 34 specialties, four difficulty levels). This benchmark, featuring expert-annotated questions, is designed for a specialized, non-English setting and sets a precedent for culturally relevant evaluation frameworks. Our open-source, scalable data pipeline, integrating a complete process from data collecting to correctness verification, facilitates the creation of this benchmark from diverse local medical resources. The benchmark and pipeline aim to foster collaboration, accelerate advancements in evaluating medical LLMs for underserved linguistic communities, and support the development of future multilingual benchmarks.

This effort holds the potential to significantly impact the field of medical AI by enabling more accurate and reliable evaluation of LLMs in languages other than English. The availability of such benchmarks will empower researchers and developers to create and refine LLMs that are better suited to the specific needs and contexts of diverse populations.  Ultimately, this can lead to improved healthcare outcomes, enhanced medical education, and more equitable access to the benefits of AI in medicine across the globe. Our data pipeline and benchmark were publicly shared for the replication and advancement of the communities.

\section{Limitations} 

Our study has several limitations. First, while our benchmark covers a wide range of medical specialties and difficulty levels, the published question format is limited to multiple-choice questions, which may not fully capture the complexities of clinical reasoning in real-world scenarios. Future work could explore incorporating other question formats, such as open-ended questions or clinical case simulations. Second, the benchmark is designed for Vietnamese, and while our data pipeline is scalable to other languages, the process of adapting and validating the benchmark for new languages may present unique challenges. These challenges include the availability of high-quality medical resources, the need for expert annotation, and the potential for cultural and linguistic variations in medical knowledge. Third, while having 100,000 entries, we were just able to verify 14,000 multiple-choice questions within 3 months. The rest of the data is valuable as it also includes clinical questions and medical records and should be exploited not just for evaluating but also for synthesizing training data. Finally, the evaluation of LLMs using our benchmark focuses primarily on their ability to answer medical questions. While this is a crucial aspect of medical LLM performance, it does not encompass other important factors such as the model's ability to generate accurate and comprehensive medical reports, engage in effective doctor-patient communication, or adhere to ethical and legal guidelines. Future research should consider a more holistic evaluation framework that assesses these additional dimensions of LLM performance in the medical domain.

\section{Acknowledge}

This work was generously supported by computing credits from Azure, AWS, and Google Cloud Platform. We also gratefully acknowledge the technical support provided by Venera AI. Additionally, we would like to express our deep gratitude to the individuals who offered invaluable support and guidance throughout the development of this project: Quynh-Anh Pham, Viet-Cuong Nguyen, and Manh-Cuong Nguyen from Thai Nguyen University of Medicine and Pharmacy; Huy-Hoang Ha, Phuc Phan, Hop Bui and Diep Duong from VietAI. 

\pagebreak

\bibliographystyle{plain}
\bibliography{refs}

\begin{thebibliography}{10}

\bibitem{abdin2024phi4technicalreport}
Marah Abdin, Jyoti Aneja, Harkirat Behl, Sébastien Bubeck, Ronen Eldan, Suriya Gunasekar, Michael Harrison, Russell~J. Hewett, Mojan Javaheripi, Piero Kauffmann, James~R. Lee, Yin~Tat Lee, Yuanzhi Li, Weishung Liu, Caio C.~T. Mendes, Anh Nguyen, Eric Price, Gustavo de~Rosa, Olli Saarikivi, Adil Salim, Shital Shah, Xin Wang, Rachel Ward, Yue Wu, Dingli Yu, Cyril Zhang, and Yi~Zhang.
\newblock Phi-4 technical report, 2024.

\bibitem{llama4}
Meta AI.
\newblock The llama 4 herd: The beginning of a new era of natively multimodal ai innovation, 2025.

\bibitem{survey2025}
Mohammed Al-Garadi, Tushar Mungle, Abdulaziz Ahmed, Abeed Sarker, Zhuqi Miao, and Michael~E. Matheny.
\newblock Large language models in healthcare, 2025.

\bibitem{claude35sonnet}
Anthropic.
\newblock Claude 3.5 sonnet, 2024.

\bibitem{LiveMedQA2017}
Asma {Ben Abacha}, Eugene Agichtein, Yuval Pinter, and Dina Demner{-}Fushman.
\newblock Overview of the medical question answering task at trec 2017 liveqa.
\newblock In {\em TREC 2017}, 2017.

\bibitem{chen2024huatuogpto1medicalcomplexreasoning}
Junying Chen, Zhenyang Cai, Ke~Ji, Xidong Wang, Wanlong Liu, Rongsheng Wang, Jianye Hou, and Benyou Wang.
\newblock Huatuogpt-o1, towards medical complex reasoning with llms, 2024.

\bibitem{chen2023meditron70bscalingmedicalpretraining}
Zeming Chen, Alejandro~Hernández Cano, Angelika Romanou, Antoine Bonnet, Kyle Matoba, Francesco Salvi, Matteo Pagliardini, Simin Fan, Andreas Köpf, Amirkeivan Mohtashami, Alexandre Sallinen, Alireza Sakhaeirad, Vinitra Swamy, Igor Krawczuk, Deniz Bayazit, Axel Marmet, Syrielle Montariol, Mary-Anne Hartley, Martin Jaggi, and Antoine Bosselut.
\newblock Meditron-70b: Scaling medical pretraining for large language models, 2023.

\bibitem{grattafiori2024llama3herdmodels}
Abhimanyu Dubey, Abhinav Jauhri, Abhinav Pandey, Abhishek Kadian, Ahmad Al-Dahle, Aiesha Letman, Akhil Mathur, Alan Schelten, Amy Yang, Angela Fan, Anirudh Goyal, Anthony Hartshorn, Aobo Yang, Archi Mitra, Archie Sravankumar, Artem Korenev, Arthur Hinsvark, Arun Rao, Aston Zhang, and Zhiwei Zhao.
\newblock The llama 3 herd of models, 07 2024.

\bibitem{gemini20}
Google.
\newblock Introducing gemini 2.0: our new ai model for the agentic era, 2024.

\bibitem{gérardin2023impacttranslationbiomedicalinformation}
Christel Gérardin, Yuhan Xiong, Perceval Wajsbürt, Fabrice Carrat, and Xavier Tannier.
\newblock Impact of translation on biomedical information extraction from real-life clinical notes, 2023.

\bibitem{hendrycks2021ethics}
Dan Hendrycks, Collin Burns, Steven Basart, Andrew Critch, Jerry Li, Dawn Song, and Jacob Steinhardt.
\newblock Aligning ai with shared human values.
\newblock {\em Proceedings of the International Conference on Learning Representations (ICLR)}, 2021.

\bibitem{hendryckstest2021}
Dan Hendrycks, Collin Burns, Steven Basart, Andy Zou, Mantas Mazeika, Dawn Song, and Jacob Steinhardt.
\newblock Measuring massive multitask language understanding.
\newblock {\em Proceedings of the International Conference on Learning Representations (ICLR)}, 2021.

\bibitem{jin2020disease}
Di~Jin, Eileen Pan, Nassim Oufattole, Wei-Hung Weng, Hanyi Fang, and Peter Szolovits.
\newblock What disease does this patient have? a large-scale open domain question answering dataset from medical exams.
\newblock {\em arXiv preprint arXiv:2009.13081}, 2020.

\bibitem{jin2019pubmedqa}
Qiao Jin, Bhuwan Dhingra, Zhengping Liu, William Cohen, and Xinghua Lu.
\newblock Pubmedqa: A dataset for biomedical research question answering.
\newblock In {\em Proceedings of the 2019 Conference on Empirical Methods in Natural Language Processing and the 9th International Joint Conference on Natural Language Processing (EMNLP-IJCNLP)}, pages 2567--2577, 2019.

\bibitem{jin2023betteraskenglishcrosslingual}
Yiqiao Jin, Mohit Chandra, Gaurav Verma, Yibo Hu, Munmun~De Choudhury, and Srijan Kumar.
\newblock Better to ask in english: Cross-lingual evaluation of large language models for healthcare queries, 2023.

\bibitem{kasai2023evaluatinggpt4chatgptjapanese}
Jungo Kasai, Yuhei Kasai, Keisuke Sakaguchi, Yutaro Yamada, and Dragomir Radev.
\newblock Evaluating gpt-4 and chatgpt on japanese medical licensing examinations, 2023.

\bibitem{kweon2024kormedmcqamultichoicequestionanswering}
Sunjun Kweon, Byungjin Choi, Gyouk Chu, Junyeong Song, Daeun Hyeon, Sujin Gan, Jueon Kim, Minkyu Kim, Rae~Woong Park, and Edward Choi.
\newblock Kormedmcqa: Multi-choice question answering benchmark for korean healthcare professional licensing examinations, 2024.

\bibitem{9184044}
Andre Lamurias, Diana Sousa, and Francisco~M. Couto.
\newblock Generating biomedical question answering corpora from q\&a forums.
\newblock {\em IEEE Access}, 8:161042--161051, 2020.

\bibitem{liu2023benchmarkinglargelanguagemodels}
Junling Liu, Peilin Zhou, Yining Hua, Dading Chong, Zhongyu Tian, Andrew Liu, Helin Wang, Chenyu You, Zhenhua Guo, Lei Zhu, and Michael~Lingzhi Li.
\newblock Benchmarking large language models on cmexam -- a comprehensive chinese medical exam dataset, 2023.

\bibitem{survey2024}
Lei Liu, Xiaoyan Yang, Junchi Lei, Yue Shen, Jian Wang, Peng Wei, Zhixuan Chu, Zhan Qin, and Kui Ren.
\newblock A survey on medical large language models: Technology, application, trustworthiness, and future directions, 2024.

\bibitem{gpt4o}
OpenAI.
\newblock Hello {GPT-4o}, 2024.

\bibitem{o3mini}
OpenAI.
\newblock Openai o3-mini pushing the frontier of cost-effective reasoning, 2025.

\bibitem{pmlr-v174-pal22a}
Ankit Pal, Logesh~Kumar Umapathi, and Malaikannan Sankarasubbu.
\newblock Medmcqa: A large-scale multi-subject multi-choice dataset for medical domain question answering.
\newblock In Gerardo Flores, George~H Chen, Tom Pollard, Joyce~C Ho, and Tristan Naumann, editors, {\em Proceedings of the Conference on Health, Inference, and Learning}, volume 174 of {\em Proceedings of Machine Learning Research}, pages 248--260. PMLR, 07--08 Apr 2022.

\bibitem{park2024assessing}
Yu~Jin Park, Aishwarya Pillai, Jing Deng, and et~al.
\newblock Assessing the research landscape and clinical utility of large language models: a scoping review.
\newblock {\em BMC Medical Informatics and Decision Making}, 24(72), 2024.

\bibitem{qin2025opportunities}
Hai Qin and Yi~Tong.
\newblock Opportunities and challenges for large language models in primary health care.
\newblock {\em J Prim Care Community Health}, 16:21501319241312571, Jan-Dec 2025.

\bibitem{qiu2024building}
Pengcheng Qiu, Chaoyi Wu, Xiaoman Zhang, Weixiong Lin, Haicheng Wang, Ya~Zhang, Yanfeng Wang, and Weidi Xie.
\newblock Towards building multilingual language model for medicine, 2024.

\bibitem{stanford_hai_medhelm}
Nigam Shah, Mike Pfeffer, and Percy Liang.
\newblock Holistic evaluation of large language models for medical applications.
\newblock Oct 2024.
\newblock Accessed: 2025-05-15.

\bibitem{singhal2022largelanguagemodelsencode}
Karan Singhal, Shekoofeh Azizi, Tao Tu, S.~Sara Mahdavi, Jason Wei, Hyung~Won Chung, Nathan Scales, Ajay Tanwani, Heather Cole-Lewis, Stephen Pfohl, Perry Payne, Martin Seneviratne, Paul Gamble, Chris Kelly, Nathaneal Scharli, Aakanksha Chowdhery, Philip Mansfield, Blaise~Aguera y~Arcas, Dale Webster, Greg~S. Corrado, Yossi Matias, Katherine Chou, Juraj Gottweis, Nenad Tomasev, Yun Liu, Alvin Rajkomar, Joelle Barral, Christopher Semturs, Alan Karthikesalingam, and Vivek Natarajan.
\newblock Large language models encode clinical knowledge, 2022.

\bibitem{gemma3}
Gemma Team, Aishwarya Kamath, Johan Ferret, Shreya Pathak, Nino Vieillard, Ramona Merhej, Sarah Perrin, Tatiana Matejovicova, Alexandre Ramé, Morgane Rivière, Louis Rouillard, Thomas Mesnard, Geoffrey Cideron, Jean-bastien Grill, Sabela Ramos, Edouard Yvinec, Michelle Casbon, Etienne Pot, Ivo Penchev, and Léonard Hussenot.
\newblock Gemma 3 technical report, 03 2025.

\bibitem{tian-etal-2019-chimed}
Yuanhe Tian, Weicheng Ma, Fei Xia, and Yan Song.
\newblock {C}hi{M}ed: A {C}hinese medical corpus for question answering.
\newblock In Dina Demner-Fushman, Kevin~Bretonnel Cohen, Sophia Ananiadou, and Junichi Tsujii, editors, {\em Proceedings of the 18th BioNLP Workshop and Shared Task}, pages 250--260, Florence, Italy, August 2019. Association for Computational Linguistics.

\bibitem{valsai}
ValsAI.
\newblock Medqa benchmark, 2025.

\bibitem{wang2025safetychallengesaimedicine}
Xiaoye Wang, Nicole~Xi Zhang, Hongyu He, Trang Nguyen, Kun-Hsing Yu, Hao Deng, Cynthia Brandt, Danielle~S. Bitterman, Ling Pan, Ching-Yu Cheng, James Zou, and Dianbo Liu.
\newblock Safety challenges of ai in medicine in the era of large language models, 2025.

\bibitem{yang2025qwen3technicalreport}
An~Yang, Anfeng Li, Baosong Yang, Beichen Zhang, Binyuan Hui, Bo~Zheng, Bowen Yu, Chang Gao, Chengen Huang, Chenxu Lv, Chujie Zheng, Dayiheng Liu, Fan Zhou, Fei Huang, Feng Hu, Hao Ge, Haoran Wei, Huan Lin, Jialong Tang, Jian Yang, Jianhong Tu, Jianwei Zhang, Jianxin Yang, Jiaxi Yang, Jing Zhou, Jingren Zhou, Junyang Lin, Kai Dang, Keqin Bao, Kexin Yang, Le~Yu, Lianghao Deng, Mei Li, Mingfeng Xue, Mingze Li, Pei Zhang, Peng Wang, Qin Zhu, Rui Men, Ruize Gao, Shixuan Liu, Shuang Luo, Tianhao Li, Tianyi Tang, Wenbiao Yin, Xingzhang Ren, Xinyu Wang, Xinyu Zhang, Xuancheng Ren, Yang Fan, Yang Su, Yichang Zhang, Yinger Zhang, Yu~Wan, Yuqiong Liu, Zekun Wang, Zeyu Cui, Zhenru Zhang, Zhipeng Zhou, and Zihan Qiu.
\newblock Qwen3 technical report, 2025.

\bibitem{yu2024tcmdtraditionalchinesemedicine}
Ping Yu, Kaitao Song, Fengchen He, Ming Chen, and Jianfeng Lu.
\newblock Tcmd: A traditional chinese medicine qa dataset for evaluating large language models, 2024.

\bibitem{ir.2024.27}
Deshiwei Zhang, Xiaojuan Xue, Peng Gao, Zhijuan Jin, Menghan Hu, Yue Wu, and Xiayang Ying.
\newblock A survey of datasets in medicine for large language models.
\newblock {\em Intelligence \& Robotics}, 4(4), 2024.

\bibitem{zhang2024ultramedical}
Kaiyan Zhang, Sihang Zeng, Ermo Hua, Ning Ding, Zhang-Ren Chen, Zhiyuan Ma, Haoxin Li, Ganqu Cui, Biqing Qi, Xuekai Zhu, Xingtai Lv, Hu~Jinfang, Zhiyuan Liu, and Bowen Zhou.
\newblock Ultramedical: Building specialized generalists in biomedicine, 2024.

\bibitem{8548603}
S.~Zhang, X.~Zhang, H.~Wang, L.~Guo, and S.~Liu.
\newblock Multi-scale attentive interaction networks for chinese medical question answer selection.
\newblock {\em IEEE Access}, 6:74061--74071, 2018.

\bibitem{zhang2024critical}
Z~Zhang and H~Ni.
\newblock Critical care studies using large language models based on electronic healthcare records: A technical note.
\newblock {\em J Intensive Med}, 5(2):137--150, 2024.

\end{thebibliography}

\clearpage
\appendix
\section{Detail of dataset}
\label{app:category_diff}
\begin{table}[h!]
\caption{Medical categories and descriptions}
\centering
\vspace{5pt}
\begin{tabular}{p{0.3\linewidth}p{0.6\linewidth}}
\toprule
\textbf{Category} & \textbf{Description} \\
\midrule
Allergy and Immunology & Diagnosis and treatment of immune system disorders, including allergies, asthma, and autoimmune diseases. \\
Anesthesiology & Anesthesia and perioperative care to manage pain and vital functions during and after surgery. \\
Cardiology & Diseases related to the heart and blood vessels. \\
Dermatology & Conditions of the skin, hair, and nails, including cosmetic issues. \\
Endocrinology & Hormonal disorders, including diabetes and thyroid diseases. \\
Gastroenterology & Disorders of the digestive system: stomach, intestines, liver. \\
Geriatrics & Health care for the elderly, disease prevention and treatment in aging adults. \\
Hematology & Blood disorders, including anemia, clotting disorders, and blood cancers. \\
Infectious Diseases & Diseases caused by bacteria, viruses, fungi, or parasites. \\
Internal Medicine & General medical care for adults across a broad range of conditions. \\
Nephrology & Kidney-related disorders, including dialysis and electrolyte issues. \\
Neurology & Disorders of the nervous system: brain, spinal cord, nerves. \\
Nuclear Medicine & Use of radioactive substances for diagnosis and treatment, especially in cancer. \\
Obstetrics and Gynecology & Women's reproductive health, pregnancy, and childbirth. \\
Oncology & Diagnosis and treatment of cancer and tumors. \\
Ophthalmology & Medical and surgical care of eye diseases. \\
Orthopedics & Musculoskeletal conditions: bones, joints, muscles. \\
Otolaryngology & Disorders of the ear, nose, throat, head, and neck. \\
Palliative Medicine & Relief from serious illness symptoms to improve quality of life. \\
Pathology & Study of disease through tissue, cell, and fluid analysis. \\
Pediatrics & Medical care for infants, children, and adolescents. \\
Physical Medicine and Rehabilitation & Restoring function for people with physical impairments or disabilities. \\
Psychiatry & Mental health, including emotional and behavioral disorders. \\
Pulmonology & Disorders of the lungs and respiratory system. \\
Radiology & Diagnostic imaging like X-rays, MRI, CT scans. \\
Rheumatology & Autoimmune and inflammatory joint/muscle diseases. \\
Sports Medicine & Injuries and prevention related to sports and exercise. \\
Surgery & Operative procedures to treat diseases and injuries. \\
Urology & Disorders of the urinary tract and male reproductive system. \\
General Medicine & Broad general care and routine health maintenance. \\
Eastern Medicine & Traditional practices: acupuncture, herbal therapy, holistic care. \\
Public Health & Population health, prevention, and policy-based approaches. \\
Preventive Healthcare & Preventing illness via vaccines, screenings, and lifestyle. \\
Emergency Medicine & Acute care for urgent and life-threatening conditions. \\
\bottomrule{}
\end{tabular}
\label{tab:category}
\end{table}
\begin{table}[h!]
\caption{Difficulty level definitions}
\centering
\begin{tabular}{p{0.1\linewidth}p{0.2\linewidth}p{0.6\linewidth}}
\toprule
\textbf{Level} & \textbf{Type} & \textbf{Description} \\
\midrule
Easy & Factual / Definitional & 
Questions about medical terms, basic definitions, or universally known concepts. \newline
\textit{Keywords:} Definitions, terminology, recall. \\
Moderate & Mechanistic / Pharmacologic & 
Focus on disease mechanisms, drug properties, or pathophysiology. \newline
\textit{Keywords:} Mechanisms, interactions, clinical application. \\
Challenging & Patient-Centered / Scenario-Based & 
Case studies requiring diagnostic reasoning or moderate clinical insight. \newline
\textit{Keywords:} Diagnostic steps, patient context, moderate complexity. \\
Hard & Advanced Clinical Reasoning & 
Complex multi-symptom cases requiring integration and critical thinking. \newline
\textit{Keywords:} Integrative reasoning, synthesis, high complexity. \\
\bottomrule
\end{tabular}
\label{tab:diff_lv}
\end{table}
\begin{figure}
    \centering
    \includegraphics[width=0.5\linewidth]{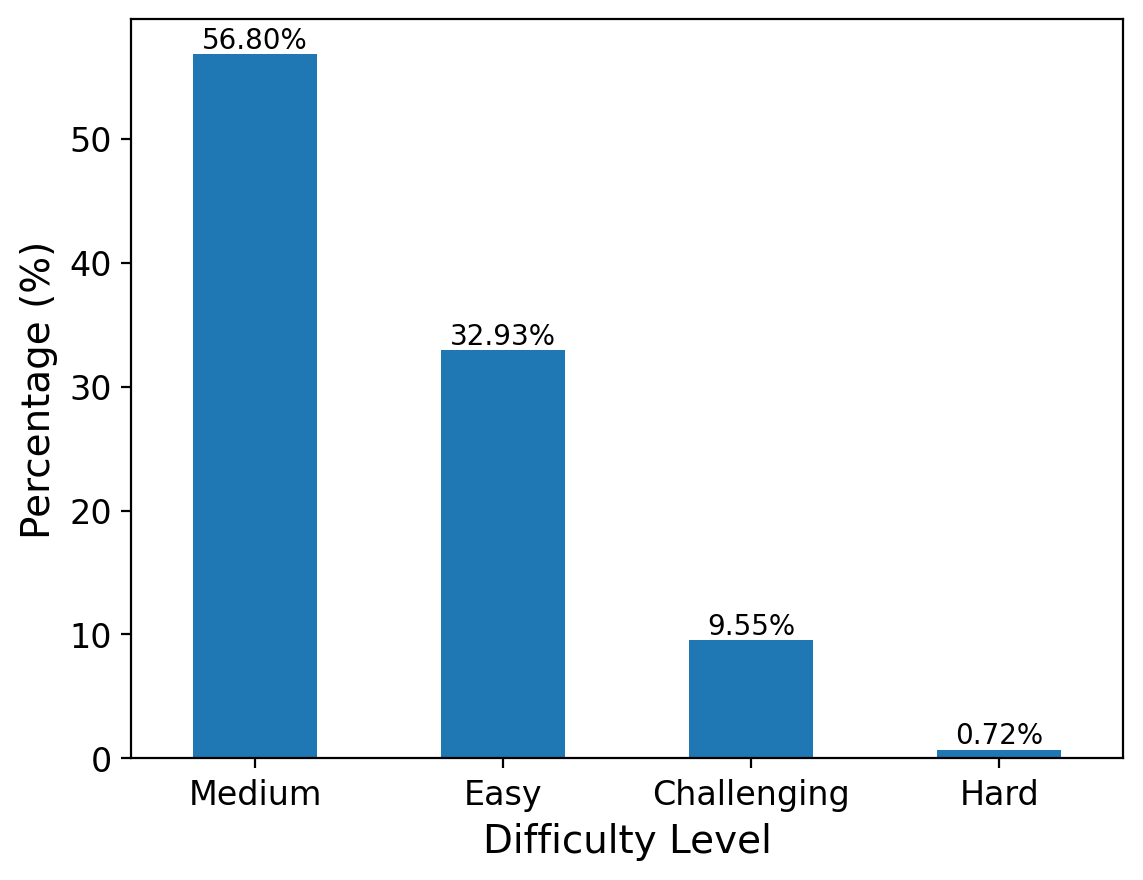}
    \caption{Difficulty level distribution of VM14K}
    \label{fig:enter-label}
\end{figure}

\section{Prompt details}
\label{app:prompt_medical}

We used the same prompt for all models:
\begin{figure}
    \centering
    \begin{verbatim}
{question}
Choose the correct option from these answers:
A. {optionA}
B. {optionB}
...


Only response with 1 character
Example:
A
    \end{verbatim}
    \caption{Prompt used in experiment}
\end{figure}

\end{document}